\documentclass{article}
\usepackage{spconf,amsmath,graphicx}

\usepackage{spconf,amsmath,graphicx,xcolor}
\usepackage{verbatim}
\usepackage{enumerate}
\usepackage{multirow}
\usepackage{graphicx}
\usepackage{subfig}
\usepackage{hyperref}

\title{SSPP-DAN: Deep domain adaptation network for \\face recognition with single sample per person}
\name{Sungeun Hong, Woobin Im, Jongbin Ryu, Hyun S. Yang}

\address{School of Computing, KAIST, Republic of Korea}

\begin{document}
%

\maketitle
\vspace*{-0.27in}
\begin{abstract}
Real-world face recognition using a single sample per person  (SSPP) is a challenging task. The problem is exacerbated if the conditions under which the gallery image and the probe set are captured are completely different. To address these issues from the perspective of domain adaptation, we introduce an SSPP domain adaptation network (SSPP-DAN). In the proposed approach, domain adaptation, feature extraction, and classification are performed jointly using a deep architecture with domain-adversarial training. However, the SSPP characteristic of one training sample per class is insufficient to train the deep architecture. To overcome this shortage, we generate synthetic images with varying poses using a 3D face model. Experimental evaluations using a realistic SSPP dataset show that deep domain adaptation and image synthesis complement each other and dramatically improve accuracy. Experiments on a benchmark dataset using the proposed approach show state-of-the-art performance. All the dataset and the source code can be found in our online repository (https://github.com/csehong/SSPP-DAN).

\end{abstract}

\vspace*{-0.05in}
\section{Introduction}
\label{sec:intro}
There are several examples of face recognition systems using a single sample per person (SSPP) in daily life, such as applications based on an ID card or e-passport \cite{lu2013discriminative}. Despite its importance in the real world, there are several unresolved issues associated with implementing systems based on SSPP. In this paper, we address two such difficulties and propose a deep domain adaptation with image synthesis to resolve these.

The first issue encountered while using SSPP is the heterogeneity of the shooting environment between the gallery and probe set \cite{xie2015blurred}. In real-world scenarios, the photo used in an ID card or e-passport is captured in a very stable environment and is often used as a gallery image. On the other hand, probe images are captured in a highly unstable environment using equipment such as surveillance cameras. The resulting image includes noise, blur, arbitrary pose, and illumination, which makes recognition difficult.

\begin{figure}[t]
\vspace*{-0.07in}
  \centering
  \subfloat[]{\includegraphics[height=6.6\baselineskip]{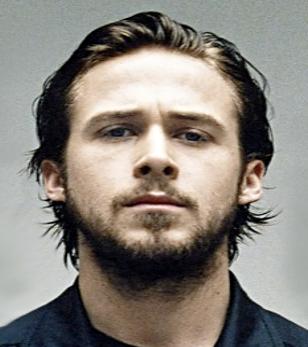}\label{fig:real1}}
    \hfill
  \subfloat[]{\includegraphics[height=6.6\baselineskip]{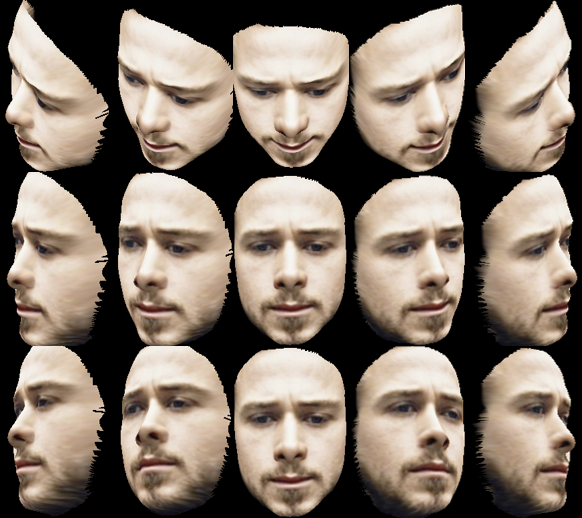}\label{fig:real2}}
  \hfill
  \subfloat[]{\includegraphics[height=6.6\baselineskip]{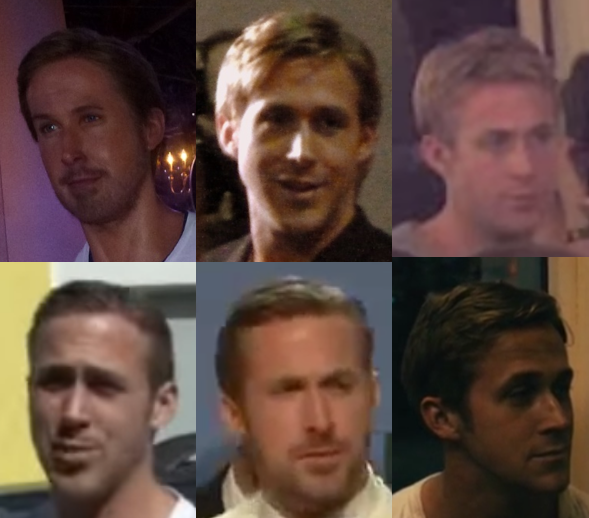}\label{fig:real3}}  
   \vspace*{-0.1in}
  \caption{Examples of (a) a stable gallery image (source domain) (b) synthetic images generated to overcome the lack of gallery samples  (source domain) (c) unstable probe images that include blur, noise, and pose variation (target domain) }
  \label{fig:real_scenario}
  \vspace*{-0.2in}
\end{figure}

\begin{figure*}
	\centering
	\includegraphics[width=0.89\linewidth]{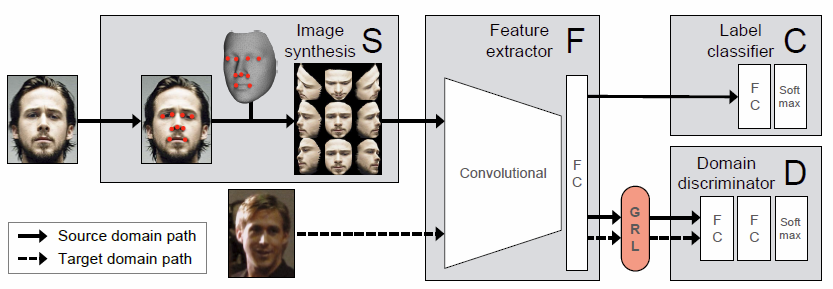}
	\vspace*{-0.08in}
	\caption{
	{
	Outline of the SSPP-DAN. Image synthesis is used to increase the number of samples in the source domain. The feature extractor and two classifiers are used to bridge the gap between source domain (i.e., stable images) and target domain (i.e., unstable images) by adversarial training with gradient reversal layer (GRL). }}
	\label{fig:overallflow}
\vspace*{-0.2in}
\end{figure*}

To address this issue, we approach SSPP face recognition from the perspective of domain adaptation (DA). Generally, in DA, a mapping between the source domain and the target domain is constructed, such that the classifier learned for the source domain can also be applied to the target domain. Inspired by this, we assume stable shooting condition of a gallery set as the source domain and unstable shooting condition of a probe set as the target domain as shown in Fig.~\ref{fig:real_scenario}. To apply DA in the unified deep architecture, we use a deep neural network with domain-adversarial training, in a manner proposed in \cite{icml2015_ganin15}. The benefit of this approach is that labels in the target domain are not required for training, i.e., the approach accommodates unsupervised learning.

The second challenge in using SSPP is in the shortage of training samples \cite{wright2009robust}. In general, the lack of training samples affects any learning system adversely, but it is more severe for deep learning approaches. To overcome this, we generate synthetic images with varying poses using a 3D face model \cite{zhang2008spacetime}  as shown in Fig.~\ref{fig:real_scenario} (center). Unlike SSPP methods that are based on external datasets \cite{wright2009robust, su2010adaptive, deng2012extended}, we generate virtual samples from an SSPP gallery set. The proposed method also differs from conventional data augmentation methods that use crop, flip, and rotation \cite{parkhi2015deep, zhang2016localize} in that it takes into account well-established techniques such as facial landmark detection and alignment that consider realistic facial geometric information. We propose a method SSPP-DAN that combines face image synthesis and deep DA network to enable realistic SSPP face recognition.

To validate the effectiveness of SSPP-DAN, we constructed a new SSPP dataset called ETRI-KAIST Labeled Faces in the Heterogeneous environment (EK-LFH). In this dataset, the gallery set was captured using a webcam in a stable environment, and the probe set was captured using surveillance cameras in an unconstrained environment. Using the experimental results, we validated that DA and image synthesis complement each other and eventually show a drastic 19.31 percentage points improvement over the baseline that does not use DA and image synthesis. Additionally, we performed experiments on the SSPP protocol of Labeled Faces in the Wild (LFW) benchmark \cite{wolf2011effective} to demonstrate the generalization ability of the proposed approach and confirmed state-of-the-art performance. 

The main contributions of this study are as follows:
({\romannumeral 1}) We propose SSPP-DAN, a method that combines face synthesis and deep architecture with domain-adversarial training.
({\romannumeral 2}) To address the lack of realistic SSPP datasets, we construct a dataset whose gallery and probe sets are obtained from very different environments. 
({\romannumeral 3}) We present a comparative analysis of the influence of DA with the face benchmark as well as with the EK-LFH dataset.

\vspace*{-0.05in}
\section{Related Works}
\label{sec:related}

A number of methods based on techniques such as image partitioning and generic learning have been proposed to address the shortage of training samples in SSPP face recognition. Image partitioning based methods augment samples by partitioning a face image into local patches \cite{lu2013discriminative, yan2014multi}. Although these techniques efficiently obtain many samples from a single subject, the geometric information of the local patch is usually ignored.  There have been attempts to use external generic sets \cite{wright2009robust, su2010adaptive, deng2012extended} by assuming that the generic set and the SSPP gallery set share some intra-class and inter-class information \cite{pei2017decision}. In this study, we augmented virtual samples from an SSPP gallery set instead of using an external set.

Several studies have proposed the application of DA for face recognition. Xie et al. \cite{xie2015blurred}  used DA and several descriptors like LBP, LPQ, and HOG  to handle the scenario in which the gallery set consists of clear images and the probe set has blurred images. Banerjee et al. \cite{banerjee2016domain} proposed a technique for surveillance face recognition using DA and a bank of eight descriptors such as Eigenfaces, Fisherfaces, Gaborfaces, FV-SIFT, and so on. Unlike the above approaches, which apply DA after extracting the handcrafted-feature from the image, we jointly perform feature learning, DA, and classification in an integrated deep architecture. Moreover, we solve the SSPP problem and consider pose variations, unlike the abovementioned approaches that only use frontal images.

A face database using surveillance camera image called SCface was proposed in \cite{grgic2011scface}. In SCface, only one person appears in each image and they are photographed at a fixed location. In contrast, the images in ours were captured in an unconstrained scenario in which 30 people were walking in the room, which induced more noise, blur, and partial occlusions.

\vspace*{-0.05in}
\section{Proposed Method}
\label{sec:method}
SSPP-DAN consists of two main components: virtual image synthesis and deep domain adaptation network (DAN) that consists of feature extractor and two classifiers. The overall flow of SSPP-DAN is illustrated in Fig.~\ref{fig:overallflow}.

\vspace*{-0.07in}
\subsection{Virtual Image Synthesis} 
\label{sec:virtual}
The basic assumption in DA is that samples are abundant in each domain and the sample distribution of each domain is similar but different (i.e., shifted from the source domain to the target domain \cite{shimodaira2000improving}). However, in the problem under consideration, there are few samples in the source domain (i.e., SSPP). In such an extreme situation, it is difficult to apply DA directly and eventually, the mechanism will fail. To address this problem, we synthesize images with changes in pose, which improves the feature distribution obtained from the face images. 

\begin{figure}[t]
  \centering
  \subfloat[
  DA fails to work because of the lack of samples in the source domain]{\includegraphics[width = 1\linewidth]{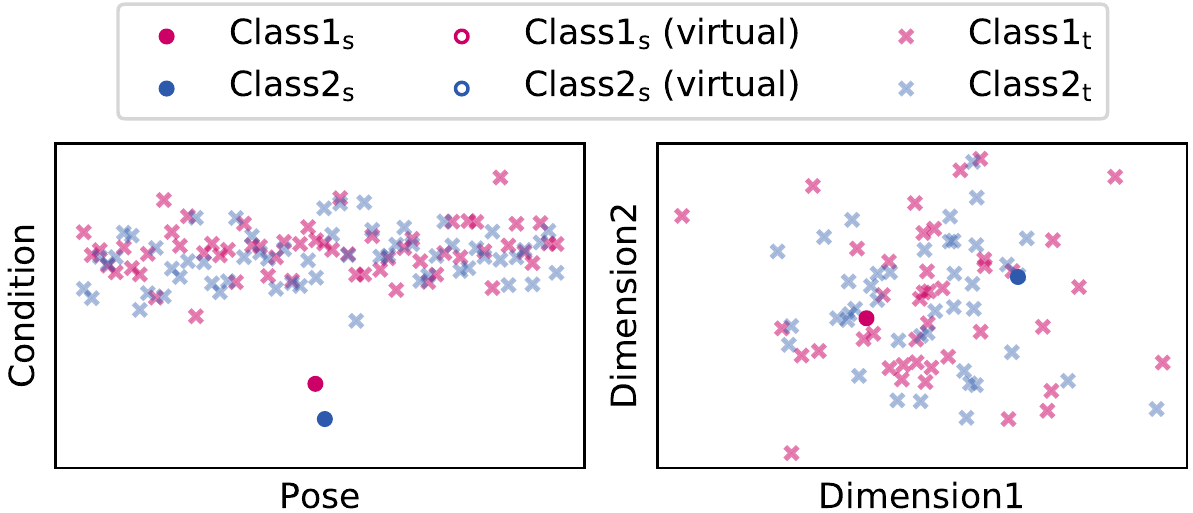}\label{fig:real1}} 
   \vspace*{-0.1in}
  \hfill
  \subfloat[
  Virtual samples along the pose axis enable successful DA, resulting in a discriminative embedding space]{\includegraphics[width = 1\linewidth]{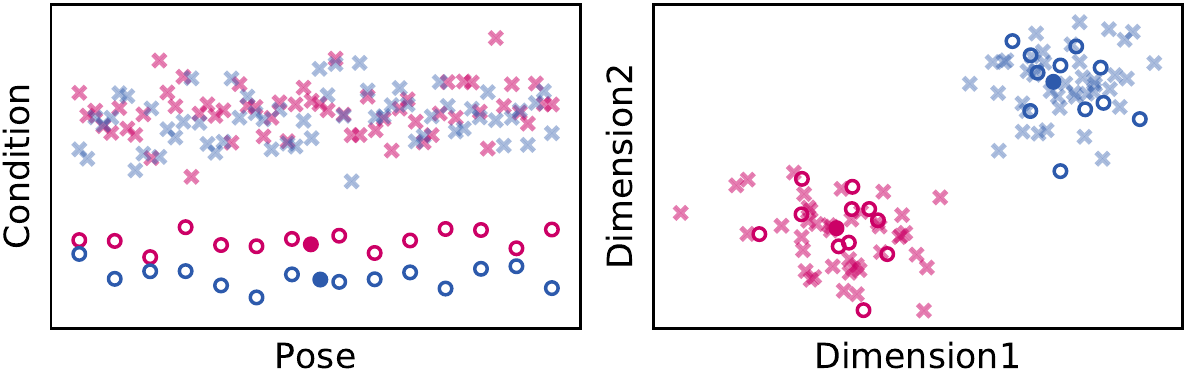}\label{fig:real2}}
  \caption{
  Facial feature space (left) and its embedding space after applying DA (right). The subscript s and t in the legend refer to the source and target domains, respectively.
  }
  \label{fig:concept_space}
  \vspace*{-0.15in}
\end{figure}

For image synthesis, we first estimate nine facial landmark points from the source domain. We use the supervised descent method (SDM) \cite{xiong2013supervised} because it is robust to illumination changes and does not require a shape model in advance. We then estimate a transformation matrix between the detected 2D facial points and the landmark points in the 3D model \cite{zhang2008spacetime, zhu2014mirror} using least-squares fit. Finally, we generate synthetic images in various poses, and these are added to the source domain as shown in Fig.~\ref{fig:concept_space}.


\vspace*{-0.07in}
\subsection{Domain Adaptation Network}
\label{ssec:DomAdaptModel}

While the variations in pose between the distributions of the two domains can be made similar by image synthesis $S$, other variations such as blur, noise, partial occlusion, and facial expression remain. To resolve the remaining differences between the two domains using DA, we use a deep network that consists of feature extractor $F$, label classifier $C$, and domain discriminator $D$. Given an input sample, it is first mapped as a feature vector through $F$. There are two branches from the feature vector\textemdash the label (identity) is predicted by $C$ and the domain (source or target) is predicted by D as shown in Fig.~\ref{fig:overallflow}.

Our aim is to learn deep features that are discriminative on the source domain during training. For this, we update the parameters of $F$ and $C$, $\theta_F$ and $\theta_C$, to minimize the label prediction loss. At the same time, we aim to train features from the labeled source domain that are discriminative in the unlabeled target domain as well (recall that  we consider unsupervised DA). To obtain the domain-invariant features, we  attempt to find a $\theta_F$ that maximizes the domain prediction loss, while simultaneously searching for parameters of $D$ ($\theta_D$)  that minimize the domain prediction loss. Taking into consideration all these aspects, we set the network loss  as
O

\vspace*{-0.15in}
\begin{equation}
	\begin{split}      
	  L &= \sum _{i\in {S}}{L}_{C}^i + \sum_{i\in {S}\cup{T}}{L}_{D}^i \:\,\,\,\qquad\textrm{when update}\,\theta_D\\	
	  L &= \sum _{i\in {S}}{L}_{C}^i - \lambda\sum_{i\in {S}\cup{T}}{L}_{D}^i  \qquad\textrm{when update}\, \theta_F, \theta_C \\	
	\end{split}
	\label{eq:loss}
\end{equation}
where ${L}_{C}^i$ and ${L}_{D}^i$ represent the loss of label prediction and domain prediction evaluated in the $i$-th sample, respectively. Here,  $S$ and $T$ denote a finite set of indexes of samples corresponding to the source and target domains.
 The parameter $\lambda$  is the most important aspect of this equation. A negative sign of $\lambda$  leads to an adversarial relationship between $F$ and $D$ in terms of loss, and its size adjusts the trade-off between them. As a result, during minimization of the network loss $L$, the parameters of F converge at a compromise point that is discriminative and satisfies domain invariance.

\vspace*{-0.05in}
\section{Experimental Results}

\vspace*{-0.07in}
\subsection{Experimental Setup}
\label{ssec:Network}
In all experiments, the face region was detected using the AdaBoost detector trained using Faces in the Wild \cite{berg2005s}. For feature learning, we fine-tuned a pre-trained CNN model, VGG-Face \cite{parkhi2015deep}, used it as the feature extractor $F$, and attached a shallow network as the label classifier $C$ (1024 - 30) and domain discriminator $D$ (1024 - 1024 - 1024 - 2). 

\vspace*{-0.07in}
\subsection{Evaluation on EK-LFH}
\label{sec:exprOurDB}

\begin{figure}[h]
	\vspace*{-0.05in}
  \centering
  \subfloat[Shooting condition for the source (left) and target (center and right)]{\includegraphics[width=\linewidth]{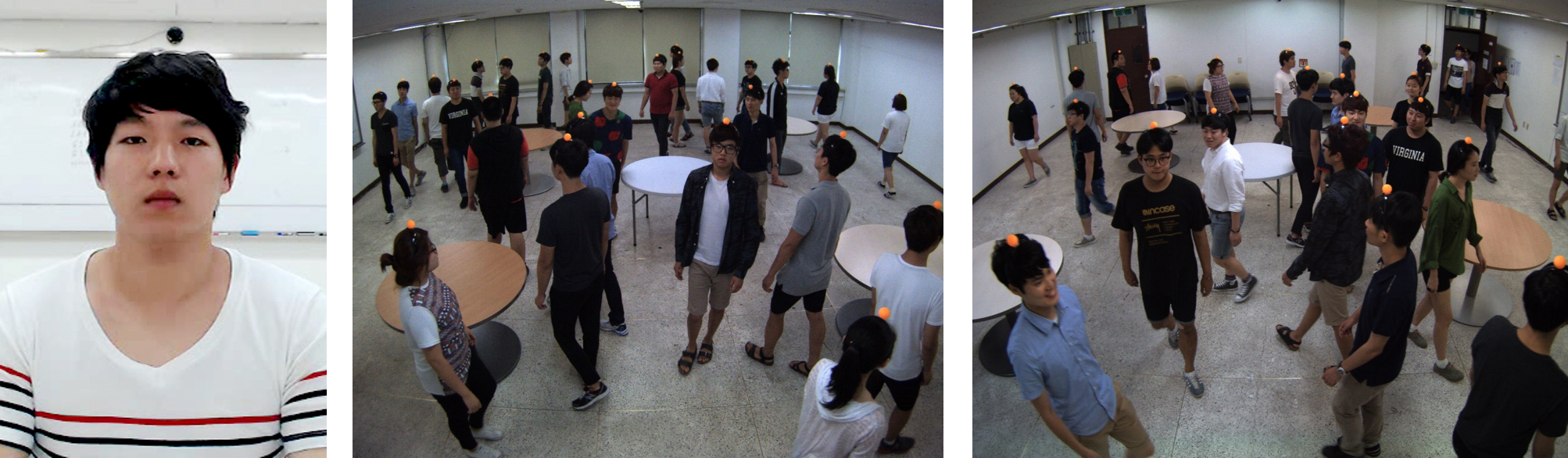}\label{fig:db1}}
    \vfill
  \vspace*{-0.1in}
  \subfloat[Face regions from the source (leftmost) and target (the others)]{\includegraphics[width=\linewidth]{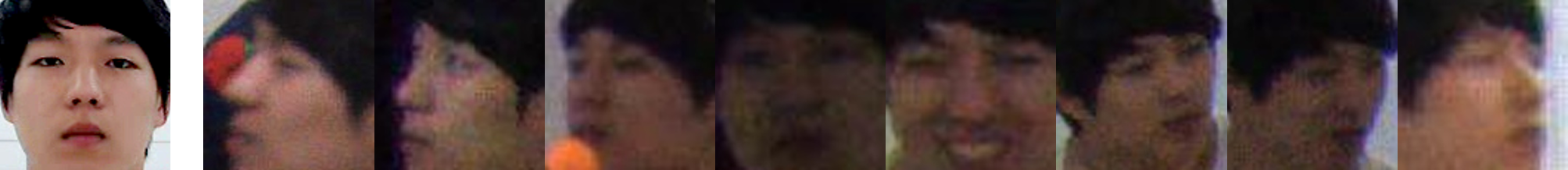}\label{fig:db2}}
    \vspace*{-0.1in}
  \caption{{Sample images in EK-LFH}  }
  \label{fig:dataset}
\end{figure}

Owing to the lack of a dataset suitable for real-world SSPP, we constructed a EK-LFH dataset containing 15,930 images of 30 subjects. Table~\ref{table:dataset_detail} shows the details of the dataset. The webcam set was used as the source domain for the training. In the surveillance set, 10,760 samples were used for training without labels in the target domain, and the rest were used for testing. Example images are shown in Fig.~\ref{fig:dataset}. 

\begin{table}[b]
\vspace*{-0.12in}
\caption{Dataset specification}
\label{table:dataset_detail}
\vspace*{-0.18in}
\begin{center}
\begin{tabular}{ l|c|c }
\hline
Domain & Source& Target \\ \hline\hline
Set  &  {webcam}  & {surveillance} \\    \hline
Subjects  &  $30$  & $30$ \\    \hline
Samples         & $30$ &  $15,900$ \\    \hline
Pose  & frontal & various \\\hline
\multirow{2}{*} {Condition}   &\multirow{2}{*} {stable}  &  unstable \\
	  &  & (blur, noise, illumination) \\   \hline
\end{tabular}
\end{center}
\vspace*{-0.05in}
\end{table}

To demonstrate the effectiveness of the proposed method, we performed evaluations using several models as shown in Table~\ref{table:result_mw} using the procedure followed in \cite{icml2015_ganin15}. The source-only model was trained using samples in the source domain, which revealed the theoretical lower bound on performance as 39.22\%. The train-on-target model was trained on the target domain with known class labels. This revealed the upper performance bound as 88.31\%. The unlabeled target domain as well as the labeled source domain were used in DAN and SSPP-DAN for unsupervised DA. Additionally, we evaluated the semi-supervised models using the same setting as DAN and SSPP-DAN, but by revealing only three labels per person in the target domain.

From Table~\ref{table:result_mw}, we clearly observe that SSPP-DAN with unsupervised as well as semi-supervised learning significantly improves accuracy. In particular, even when the labels of the target domain are not given, the accuracy of the proposed SSPP-DAN was 19.31 percentage points higher than that for source-only. The fourth and fifth rows validate the importance of image synthesis when applying unsupervised DA. Adding synthesized virtual images to the training set increased the performance by 27.42 percentage points. Interestingly, as shown in the third row, adding synthetic images to source-only degrades performance. This result indicates that image synthesis alone cannot solve the SSPP problem efficiently, instead DA and image synthesis operate complementarily  in addressing the SSPP problem.

\begin{table}
\caption{Recognition rates (\%) for different models and different training sets of the EK-LFH
}
\label{table:result_mw}
\vspace*{-0.18in}
\begin{center}
\begin{tabular}{l|l|c }
\hline
Model & Training set & Accuracy \\\hline
\hline
\multirow{2}{*} {Source only}   & $\textrm{S}$ &  39.22  \\
	  & $\textrm{S} + \textrm{S}_\textrm{v}$ & 37.15 \\                  
\hline
DAN     &$\textrm{S} + \textrm{T}$ &  31.11   \\ 
\textbf{SSPP-DAN}	& $\textrm{S} + \textrm{S}_\textrm{v} + \textrm{T}$ & \textbf{58.53} \\  
\hline
{Semi DAN}   & $\textrm{S} + \textrm{T} + \textrm{T}_\textrm{l}$  &  67.28   \\
\textbf{Semi SSPP-DAN}    &$\textrm{S} + \textrm{S}_\textrm{v} + \textrm{T} + \textrm{T}_\textrm{l}$ & \textbf{72.08} \\  
\hline
Train on target	  & $\textrm{T}_\textrm{l}$ & 88.31 \\     
\hline
\multicolumn{3}{c}{$\textrm{S}$: Labeled webcam \quad $\textrm{T}$: Unlabeled surveillance} \\
\multicolumn{3}{c}{$\textrm{S}_\textrm{v}$: Virtual set from $\textrm{S}$ \quad $\textrm{T}_\textrm{l}$: Labeled surveillance} \\
\end{tabular}
\end{center}
\vspace*{-0.3in}
\end{table}

\vspace*{-0.07in}
\subsection{Evaluation on LFW for SSPP}
\label{sec:exprBenchmark}
In order to demonstrate the generalization ability of SSPP-DAN, we performed an additional experiment on the LFW using the proposed SSPP method. For fair comparison with previous SSPP methods, we used LFW-a \cite{wolf2011effective}, and followed the experimental setup described in \cite{yang2016joint}. The LFW for SSPP included images from 158 subjects, each of which contained more than 10 samples, as well as the labels of all subjects. The first 50 subjects were used as probe and gallery, and the images of the remaining 108 subjects were used as a generic set. For the 50 subjects, the first image was used as the gallery set and the remaining images were used as the probe set.

Since the LFW did not consider DA originally, it made no distinction between source and target domain. Hence, we used the original generic set as the source domain and the synthetic images from the generic set as the target domain. We applied DA in a supervised manner to generate a discriminative embedding space. After training, we used the output of the last FC layer as the feature, and implemented prediction using the linear SVM. We also evaluated fine-tuned VGG-Face without image synthesis and DA. 
Experiments using the benchmark confirmed that VGG-face based methods including ours have superior discriminative power over other approaches as shown in Table~\ref{table:lfwResult}.
This indicates the generality of deep features from the VGG-Face trained on a large scale dataset.
It is apparent from this table that, by comparing VGG-Face with the proposed method, the combination of image synthesis and DA shows promising results in the `wild' dataset.

\begin{table}
\caption{Recognition rates (\%) on LFW dataset for SSPP}
\label{table:lfwResult}
\vspace*{-0.18in}
\begin{center}
\begin{tabular}{l|c|l|c}
\hline
Method & Accuracy  & Method & Accuracy  \\\hline
\hline
DMMA \cite{lu2013discriminative}& 17.8 & RPR \cite{gao2015neither} & 33.1   \\ 
AGL \cite{su2010adaptive} & 19.2 & DeepID \cite{sun2014deep} & 70.7 \\
SRC \cite{wright2009robust} & 20.4 & JCR-ACF \cite{yang2016joint} & 86.0 \\
ESRC \cite{deng2012extended} & 27.3 &  VGG-Face \cite{parkhi2015deep}& 96.43  \\
LGR \cite{zhu2014local} & 30.4 & \textbf{Ours} & \textbf{97.91} \\
\hline
\end{tabular}
\end{center}
\vspace*{-0.18in}
\end{table}

\vspace*{-0.05in}
\section{CONCLUSION}
\label{sec:con}
This paper proposed a method based on integrated domain adaptation and image synthesis for SSPP face recognition, especially for cases in which the shooting conditions for the gallery image and the probe set are completely different.  Synthetic images generated in various poses were used to deal with the lack of samples in the SSPP. In addition, a deep architecture with domain-adversarial training was used to perform domain adaptation, feature extraction, and classification jointly. Experimental evaluations showed that the proposed SSPP-DAN had an accuracy 19.31 percentage points higher than that of the source-only baseline even when the labels of the target domain were not given. Our method also achieved state-of-the-art results on the challenging LFW for SSPP. In future work, we plan to expand our approach to a fully trainable architecture including image synthesis as well as domain adaptation using standard back-propagation.

\bibliographystyle{IEEEbib}
\bibliography{refs}

\end{document}